\documentclass[preprint,12pt,3p]{elsarticle}

\usepackage{lineno}

\usepackage[utf8]{inputenc}
\usepackage{soul,color}
\usepackage{paralist}

\usepackage[numbers]{natbib}
\usepackage{mathtools}
\usepackage{amssymb}
\usepackage{amsfonts}
\usepackage{amsmath}
\usepackage{lipsum}
\usepackage{graphicx}
\usepackage{epstopdf}
\usepackage{hyperref}
\usepackage[export]{adjustbox}
\usepackage{float}
\usepackage{tabularx}
\usepackage{multirow}
\usepackage{rotating}
\usepackage[super]{nth}

\DeclareMathOperator*{\argmin}{arg\,min}

\newcommand{\bs}[1]{\boldsymbol{#1}}

\newcommand{\mc}[1]{\mathcal{#1}}

\newcommand{\grad}[2]{\frac{\partial #1}{\partial #2}}

\newcommand{\Ce}[0]{\boldsymbol{\mathcal{C}}}

\newcommand{\sig}[0]{\boldsymbol{\sigma}}
\newcommand{\dsig}[0]{\dot{\boldsymbol{\sigma}}}

\newcommand{\eps}[0]{\boldsymbol{\varepsilon}}
\newcommand{\deps}[0]{\dot{\boldsymbol{\varepsilon}}}
\newcommand{\epsd}[0]{\boldsymbol{\varepsilon}_d}
\newcommand{\depsd}[0]{\dot{\boldsymbol{\varepsilon}}_d}

\newcommand{\dg}[0]{\dot{\gamma}}
\newcommand{\esig}[0]{\tilde{\sig}}

\newcommand{\dyad}[0]{\otimes}

\usepackage[capitalise]{cleveref}

\journal{Elsevier}

\begin{document}

\begin{frontmatter}

\title{Constitutive model characterization and discovery \\
using physics-informed deep learning}

\author[UBC]{
    \href{https://orcid.org/0000-0003-2659-0507}{\includegraphics[scale=0.06]{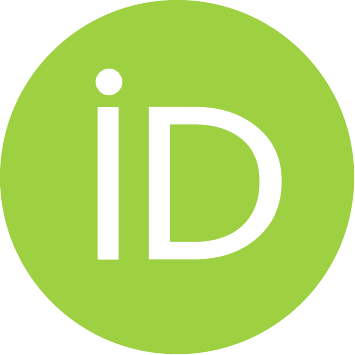}\hspace{1mm}Ehsan Haghighat}\corref{cor1}}
\address[UBC]{Department of Civil Engineering, University of British Columbia, Vancouver, BC, Canada}
\cortext[cor1]{Correspondence to: Ehsan Haghighat}
\ead{ehsan.haghighat@ubc.ca}

\author[UBC]{Sahar Abouali}
\ead{sahar@composites.ubc.ca}

\author[UBC]{
    \href{https://orcid.org/0000-0001-5101-0661}{\includegraphics[scale=0.06]{orcid.pdf}\hspace{1mm}Reza Vaziri}
}
\ead{reza.vaziri@ubc.ca}

\begin{abstract}
Constitutive models are fundamental blocks of modeling physical processes, where they connect conservation laws with the kinematics of the system. 
They are often expressed in the form of linear or nonlinear systems of ordinary differential equations (ODEs). Within nonlinear regimes, however, it is often challenging to characterize these constitutive models. 
For solids and geomaterials, the constitutive relations that relate the macroscopic stress and strain quantities are described using highly nonlinear, constrained ODEs to characterize their mechanical response at different stages of both reversible and irreversible deformation process. 
A recent trend in constitutive modeling leverages complex neural network architectures to construct \emph{model-free} material models, however, such complex networks are inefficient and demand significant training data. 
Therefore, we believe theory-based parametric models of elastoplasticity are still the most efficient and predictive. To alleviate the challenging task of characterization and discovery of such models, here, we present a physics-informed neural network (PINN) formulation for stress-strain constitutive modeling. 
The main obstacle that we address is to have complex inequality constraints of elastoplasticity theory embedded in the PINN loss functions. These constraints are crucial to find the correct form of the yield surface and plastic flow. We also show that calibration of new datasets can be performed very efficiently 
and that enhanced performance can be achieved even
for the case of discovery. This framework requires a single dataset for characterization. 
Although we only focus on mechanical constitutive models, similar analogies can be used to characterize constitutive models for any physical process.
\end{abstract}


\end{frontmatter}


\section{Introduction}
Whether for heat transfer in solids or fluid flow in porous media or deformation of continua, constitutive models are essential components of a mathematical description for various physical processes. They connect conservation laws with kinematics of the system. For instance, Fourier law expresses a macroscopic view of thermal energy transfer in a system due to collision (kinetic energy) of its molecules. 

For solids and geomaterials, mechanical constitutive models (material laws) express a macroscopic view of force and displacement correspondence, resulting from multiscale mechanisms such as molecular deformation, granular displacements, or mesoscale deformation localization, in terms of homogenized (averaged) quantities, i.e., stresses and strains, in a representative elementary volume (REV) \cite{Nemat-Nasser1999Micromechanics:Materials}.
They build the foundation to correlate conservation laws, e.g., conservation of momentum, with the kinematics of the system, e.g., displacements. They are predictive and in their discretized form, they are also computationally efficient as they result in relatively simple algebraic expressions between stress and strain increments. 
They are therefore the backbone of advanced analysis and design frameworks, such as the Finite Element Method (FEM), and have been used extensively for analysis and design of engineering infrastructures such as skyscrapers, dams, nuclear power plants, aircrafts and space shuttles \cite{Zienkiewicz2013TheFundamentals,Zienkiewicz2014TheMechanics, Belytschko2000NonlinearStructures}.
However, they are phenomenological in nature, derived empirically for a specific material based on experimental observations, and it is not straightforward to calibrate or extend them for new materials. 
Among many theories developed over the past century, including hyperelasticity, viscoelasticity, elastoplasticity, hypoplasticity, and damage mechanics, elastoplasticity and damage theories still remain popular, in both research and application, and therefore are the focus of this manuscript \cite{Lubliner1992PlasticityTheory, Simo1997, deSouzaNeto2008ComputationalPlasticity,Pietruszczak2010FundamentalsGeomechanics, Borja2013Plasticity:Computation}. 

While constitutive models are designed to express the mechanical response of materials at a REV scale, they can be derived in different ways, by testing at REV scale, by testing at larger scales, or by homogenizing microscale or even molecular-scale simulations. Therefore, there are two types of macroscopic mechanical testing on materials, each designed to study certain behaviors: 
\begin{itemize}
    \item \textit{Homogeneous (REV) test} where the state of stress and strain remains homogeneous within the samples and the problem is described as a point or unit element. This test is suitable for ductile materials exhibiting elastoplastic response. Uniaxial or triaxial tests on metallic coupons or cubic/cylindrical specimens of  geomaterials belong to this class of problems.  
    \item \textit{Full-field test} where the state of the stress and strain is not homogeneous and the problem is described as a boundary value problem (BVP). This test is often set up to study fast processes such as cracking in brittle materials. Three-point or four-point bending tests on concrete beams fall into this category. 
\end{itemize}
In this paper, we are focused on the first class of problems, i.e., calibration of constitutive models derived from homogeneous testing.

Elastoplastic constitutive models have been developed extensively to describe the mechanical response of different materials, e.g. metals or granular materials, under various loading conditions and with isotropic or anisotropic considerations. They often result in a set of nonlinear ordinary differential equations (ODEs). 
They can efficiently and accurately predict the history-dependent response of material by leveraging (i) a \emph{yield surface}, an evolving surface defining the transition between reversible (elastic) and irreversible (elastoplastic) deformations, (ii) a \emph{plastic flow rule}, defining the direction of irreversible deformation, and (iii) a \emph{kinematic rule}, defining the decomposition law of the total strain into reversible and irreversible components \cite{Simo1997}. These models work efficiently with numerical solvers such as FEM and have been employed commonly to analyze complex structural and mechanical systems. The yield function or the flow rule, however, are defined empirically by performing extensive uniaxial or multiaxial or even full-field (full-scale) experiments for each material. A major challenge that has been an active area of research is the development of algorithms for calibration and extension of these models to new materials and tests, which we also attempt to address here. 

Driven by the success of deep learning techniques in various areas \cite{NIPS2012_4824,graves2013speech,Lecun2015,Goodfellow2016}, a recent trend in constitutive modeling suggests replacing material laws with complex neural networks. Although the origins of this framework date back to the 1990s \cite{Ghaboussi1991KnowledgeBasedNetworks, Ghaboussi1998NewModeling, Hashash2004NumericalAnalysis}, the recent advancements in neural network architectures \cite{Goodfellow2016} and availability of advanced frameworks such as TensorFlow have reignited this trend. 
In their state of the art, these models can accurately predict the history-dependent mechanical response of different materials under loading and unloading cycles \cite{Kirchdoerfer2016Data-drivenMechanics,Mozaffar2019,Jang2021MachinePlasticity,Karapiperis2021Data-DrivenMechanics}. 
These methods, however, come with several drawbacks. They are understood as "black-box" models, and they demand a significant amount of training data, which is not desirable for most engineering applications. 
Additionally, due to their number of parameters and complex network architecture, their computational performance is questionable when it comes to coupling them with FEM or other numerical methods for real-scale applications. Lastly, due to their interpolative nature, these models cannot be used confidently outside the training set. 

These drawbacks suggest that the rational nature of plasticity theories, given their remarkable accuracy and performance with a limited number of parameters, are still better choices for designing material laws that can be used for large-scale analysis. 
However, as discussed, it is a complex thought process to come up with new advanced models. Then the question is whether we can keep the theory intact and replace this complexity, i.e. calibration or discovery, with modern and automated artificial intelligence (AI) frameworks. This is indeed an active topic of research with a very limited number of available studies. However, given the success of algorithms such as sparse regression methods \cite{Brunton2016, Rudy2019} or physics-informed neural networks \cite{Raissi2019, Karniadakis2021Physics-informedLearning}, we can foresee that ultimately we will have AI algorithms that can perform REV constitutive calibration and discovery tasks reliably. In fact, \citet{Flaschel2021UnsupervisedLaws} recently proposed the use of sparse regression techniques for discovering hyperelastic constitutive models. Here, we formulate the calibration and discovery problem using physics-informed neural networks. 

Classical calibration (inversion) methods used in the context of constitutive models are mostly based on least squares optimization \cite{Najjar1990Elasto-plasticEvaluation,Anandarajah1991Computer-aidedModel,Sol1997MaterialMethods, Ghouati1998IdentificationProcesses,Zentar2001IdentificationAnalysis,Cekerevac2006CalibrationProcedure,Cooreman2007ElastoplasticMatrix,Zhang2013ParameterOptimization, Liu2016DeterminingDatabase,Haghighat2020AShale}. Accordingly, an objective function is defined to assess the error between the output of the numerical model and its expected experimental values. The challenge with such optimization models is that for every parameter update, one needs to re-run the forward numerical model to re-evaluate the error function. Additionally, some optimization methods, such as those from the gradient descent family, need the gradient vector evaluated using expensive finite differentiation. An alternative approach is to construct an interpolation model, e.g., in the form of neural networks, that intakes a reduced form of the experimental data, i.e., force-displacement or stress-strain data, and outputs their corresponding material parameters \cite{Gangopadhyay1999SubsurfaceGIS,Huber2000DeterminationNetworks,Obrzud2009OptimizationNetworks,Kulga2018CharacterizationNetwork,Yang2019UsingLimit,Zobeiry2020Theory-guidedComposites}. Once trained, these models are extremely fast in performing inference. However, they require a significant amount of training data, often obtained by a brute-force search in the admissible space of parameters, and therefore they suffer the curse of dimensionality. They remain good candidates for industrial setups where repeated experimentation is needed on similar materials. 

Based on the foregoing discussion, the most recent trend is the development of explainable AI models that can facilitate model calibration and discovery with minimal data \cite{Vlassis2021SobolevHardening,Sun2021Data-drivenPropagation,Flaschel2021UnsupervisedLaws,Huang2020LearningNetworks,Xu2021LearningNetworksb}. In this manuscript, we propose a novel approach using Physics-Informed Neural Networks (PINN). Introduced recently by Raissi et al. \cite{Raissi2019}, PINNs have been an active area of research in the last few years and have been applied to forward and inverse solutions of various problems in fluid mechanics\cite{Raissi2019,Jin2021NSFnetsEquations,Cai2021Physics-informedReview,Reyes2020LearningFluids}, solid mechanics \cite{Haghighat2021AMechanics, Rao2021Physics-InformedData, Haghighat2021AOperator,Guo2020AnElasticity}, heat transfer \cite{Cai2021Physics-informedReview,AminiNiaki2021Physics-informedManufacture}, and flow and transport in porous media \cite{Fuks2020LimitationsMedia,Bekele2021Physics-informedConsolidation,Haghighat2021Physics-informedTraining}, to name a few. 
In all these studies, PINNs have been used in the context of solving an initial or boundary value problem (BVP), and some obstacles have been found to consider highly nonlinear material models. 

In this study, we focus primarily on the constitutive modeling itself, an idealization that allow us to consider far more complex constitutive models than previously studied in the context of PINNs. We formulate a PINN elastoplasticity solver (loss functions) by leveraging the constraints of elastoplasticity and damage theories to arrive at explainable constitutive models given standard stress-strain data. We validate the proposed framework for calibration of von Mises elastoplasticity model with isotropic hardening, kinematic hardening, mixed hardening, pressure-dependent Drucker-Prager model, and with coupled damage-plasticity model \cite{Mises1913MechanikZustand, Drucker1952SoilDesign, Lubliner1992PlasticityTheory}. Our main contributions include:
\begin{enumerate}
\item[i.]  Formulating inequality constraints of elastoplasticity as PINN constraints; 
\item[ii.] Leveraging transfer learning to perform material characterization (calibration) very efficiently;
\item[iii.] Applying the framework to various models of elastoplasticity from the von Mises family.
\end{enumerate} Therefore the framework is capable of characterizing and discovering yield surface and flow rules of elastoplasticity. 

Here, we use PINNs in the context of single-element material model calibration where stress and strain states can be considered to be homogeneous. Therefore, it applies to uni-axial or multi-axial experimental setups. It is also applicable to data obtained after homogenizing stress-strain data from meso- or micro-mechanical or even molecular dynamic simulations.  For full-scale experiments, one may leverage the boundary-value-problem setup of PINNs \cite{Haghighat2021AMechanics}. 
Note that one can also formulate the problem of constitutive model characterization using sparse regression techniques \cite{Brunton2016,Flaschel2021UnsupervisedLaws}. The drawback would be that it would require an independent forward solver for loss evaluation that can be costly. However, their advantage would be that sparse-regression methods may reach higher accuracy by decoupling calibration and forward solver and by enforcing sparsity.

\section{Constitutive models} \label{sec:Constitutive}

This section summarizes basics of the incremental theory of elastoplasticity with isotropic damage (see \cite{Lubliner1992PlasticityTheory, Simo1997} for additional details).

\subsection{Elastoplasticity}
According to the theory of elastoplasticity, the state of a material as a function of Cauchy stress state $\bs{\sigma}$ and hardening parameter $\bs{q}$ is described using a yield surface $\mc{F}$. The elastic domain $\mathbb{E}_\sigma$ and yield surface $\partial\mathbb{E}_\sigma$ are defined as

\begin{align}
    \mathbb{E}_\sigma := \{(\sig, \bs{q}) | \mc{F}(\sig, \bs{q}) < 0 \} \label{eq1}, \\
    \partial\mathbb{E}_\sigma := \{(\sig, \bs{q}) | \mc{F}(\sig, \bs{q}) = 0 \}. \label{eq2} 
\end{align}

\noindent
The normal vector to the yield surface $\partial\mathbb{E}_\sigma$ is then expressed as 

\begin{align}
\bs{n} &= \grad{\mc{F}(\sig,\bs{q})}{ \sig} \label{eq3}.
\end{align}

\begin{figure}
    \centering
    \includegraphics[width=1\textwidth]{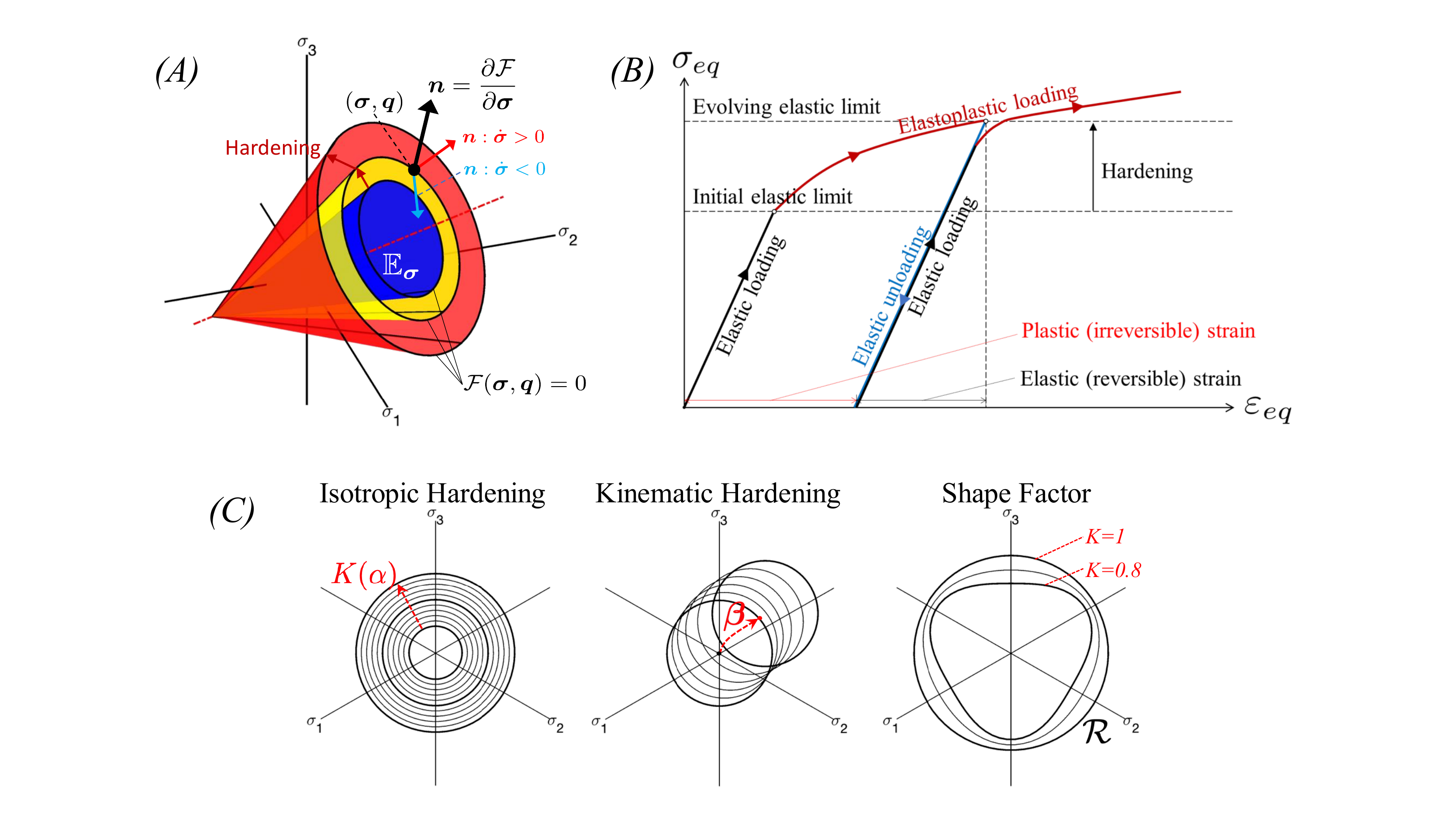}
    \caption{(A) A schematic plot of evolving yield surface $\mc{F}(\sig, \bs{q})$, with normal vector $\bs{n}$, and elastic domain $\mathbb{E}_{\sig}$. The elastoplastic loading from an arbitrary state $(\sig, \bs{q})$ on the yield surface is shown with the red-arrow ($\bs{n}:\dot{\sig}>0$), while the elastic unloading is depicted with the blue arrow ($\bs{n}:\dot{\sig}<0$). (B) A schematic plot of different loading stages in equivalent stress and strain coordinates. (C) Isotropic and kinematic hardening rules as well as the yield surface shape factor $\mc{R}$ are plotted on the $\pi$-plane (a plane, perpendicular to the hydrostatic axis $\sigma_1=\sigma_2=\sigma_3$). For the Drucker-Prager family, the shape factor $\mc{R}$ is taken as $\mc{R} = \frac{1}{2}\left[1 + 1/K - (1-1/K)(q/r)^3\right]$. } 
    \label{fig1}
\end{figure}

\noindent
This is shown schematically in \Cref{fig1}-A. 

Within the elastic domain $\mathbb{E}_{\sigma}$, the material's response remains elastic, and based on the Hooke's law, the stress increment is expressed as $\dsig = \Ce : \deps$, where $\deps$ is the total strain rate, and $\Ce$ is the elastic stiffness operator, defined as

\begin{align}
\Ce = (\kappa-\frac{2}{3}\mu) \bs{1}\dyad\bs{1} + 2\mu\mathbb{I} \label{eq4}
\end{align}

\noindent
where $\bs{1}$ and $\mathbb{I}$ are \nth{2} and \nth{4} order identity tensors, $\kappa$ is the bulk modulus and $\mu$ is the shear modulus. Equivalently, we can also express \cref{eq4} as a function of elastic modulus $E$ and Poisson's ratio $\nu$ since $\mu=E/2(1+\nu)$ and $\kappa=E/3(1-2\nu)$. On the evolving yield surface $\partial \mathbb{E}_{\sigma}$, the material behaves elastoplastically and the total strain rate is decomposed additively into elastic and plastic parts, i.e., $\deps = \deps^e + \deps^p$. The constitutive relation for the stress increment is then expressed as 

\begin{equation}\label{eq5}
    \dsig = \Ce : \deps^e = \Ce : (\deps - \deps^p).
\end{equation}

The flow rule formulation of plasticity hypothesizes that the plastic deformation occurs in a direction $\bs{r}$ normal to a potential surface $\mc{G}$: 

\begin{align}
\bs{r} &= \grad{\mc{G}(\sig, \bs{q})}{ \sig}, \label{eq6} \\
\deps^p &= \dg\bs{r}(\sig, \bs{q}), \label{eq7} \\
\dot{\bs{q}} &= \dg\bs{h}(\sig, \bs{q}). \label{eq8}
\end{align}

\noindent
Here, $\bs{r}$ indicates the plastic flow direction, and $\bs{h}$ expresses the hardening rule. The plastic multiplier $\gamma$ is defined as $\gamma(t) = \int_{t_0}^{t} \dg dt$. The internal variable $\bs{q}$ traces the hardening in the material as a function of $\{\eps^p, \sig\}$, and is subdivided into an expansion tracer $\bs{\alpha}$ and a relocation tracer $\bs{\beta}$, known as isotropic and kinematic hardening, respectively, as $\bs{q}=[\bs{\alpha}, \bs{\beta}]^T$. If $\mc{G} = \mc{F}$, the flow rule is called associative, otherwise it is non-associative. 

The isotropic hardening parameter $\bs{\alpha}$ is often taken as either volumetric or deviatoric invariants of the plastic strain tensor, i.e., $\bs{\alpha} \in \{\varepsilon_v^p,\varepsilon_q^p\}$. The kinematic hardening tensor traces the center of the yield surface and is evaluated as $\dot{\bs{\beta}} = \dg \bs{h}_{\beta}$. The hardening variable $\bs{q}$ is then evaluated through the integration of the hardening rule as $\bs{q}=\int_0^t \dg \bs{h} dt$. Such models are further detailed graphically in \Cref{fig1}.

Based on the current state of stress tensor $\sig$ and yield surface $\mc{F}(\sig, \bs{q})$, the plastic multiplier $\dot{\gamma}$ can be categorized as follows: 

\begin{equation*}\label{eq:plas-loading-unloading}
\begin{split}
\mc{F}<0 &: \quad \dg = 0 ~~~\textrm{(Elastic Region)}\\
\bs{n}:\dsig < 0 \text{~~and~~} \mc{F}=0  &:  \quad \dg = 0 ~~~\textrm{(Elastic Unloading)}\\
\bs{n}:\dsig = 0 \text{~~and~~} \mc{F}=0  &: \quad \dg = 0 ~~~\textrm{(Neutral Loading)}\\
\bs{n}:\dsig > 0 \text{~~and~~} \mc{F} = 0 &: \quad \dg > 0 ~~~\textrm{(Plastic Loading)}\\
\mc{F} > 0 &: \quad \textrm{Invalid State of Stress}
\end{split}
\end{equation*}

\noindent
In summary, they form the following conditions, also known as Karush–Kuhn–Tucker (KKT) conditions \cite{Simo1997}:

\begin{align}
    \mc{F}\dg = 0, \quad \mc{F}\le0, \quad \dg\ge0, \label{eq9}
\end{align}

\noindent
which also implies the consistency condition:

\begin{equation}\label{eq10}
	\dot{\mc{F}} = \frac{\partial \mc{F}}{\partial \sig}:\dsig + \frac{\partial \mc{F}}{\partial \bs{q}}\cdot{\bs{h}}~\dg=0.
\end{equation}

\noindent
Substituting \cref{eq5} back into \cref{eq10} and after some algebraic manipulations, we can arrive at the following relation for the plastic multiplier:

\begin{equation}\label{eq11}
    \dg = \frac{\bs{n}:\Ce:\deps}{\bs{n}:\Ce:\bs{r} - \frac{\partial \mc{F}}{\partial \bs{q}}\cdot\bs{h}}.
\end{equation}

\subsection{Coupled damage elastoplasticity}
Continuum damage mechanics (CDM) is a popular and computationally efficient approach for modeling  progressive failure of materials (\cite{kachanov1986introduction, lemaitre1985continuous,krajcinovic1987continuum}). 
Based on the CDM theory, the effect of local material failure in the form of micro-cracks is expressed by its elastic stiffness degradation using a damage variable   $\omega$, which results into a strain-softening behavior. 
The CDM formulation can be coupled with plasticity theories to account for both plastic deformation and damage in the material  \cite{kachanov1958rupture}. In a coupled damage elastoplastic material model involving the commonly used strain equivalance approach, the constitutive relation can be formulated in terms of the effective stress state $\tilde{\bs{\sigma}}$ \cite{lemaitre1985continuous}, as

\begin{align}
\esig = \frac{\sig}{1-\omega}, \label{eq12}
\end{align}

\noindent
where, $\sig$ is the Cauchy stress tensor representing the actual strate of stress in the damaged material and $\esig$ is the effective stress tensor representing the state of stress in an equivalent undamaged material subject to the same strain state. $\omega$ denotes the isotropic scalar damage variable, which varies between 0 and 1. For any given cross-section, the factor $1-\omega$ expresses the ratio of the effective load-carrying area with respect to the overall cross-sectional area. 

The yield and potential surfaces are defined based on the effective stress tensor $\mc{F}(\esig, \tilde{\bs{q}})$ and $\mc{G}(\esig, \tilde{\bs{q}})$, and the hardening parameter is expressed as $\bs{h}(\esig, \tilde{\bs{q}})$. In general, the damage variable $\omega$ is a function of the hardening parameter in the effective space, i.e., $\omega=\omega(\tilde{\bs{q}})$. The rate of change of the Cauchy stress tensor can then be written as 

\begin{align}
\dot{\sig} = (1-\omega) \Ce:(\deps - \deps^p) - \dot{\omega}\Ce:(\eps - \eps^p). \label{eq13}
\end{align}

\noindent
Here, the plastic flow rule, hardening rules, KKT conditions and plastic multiplier are similarly formulated to those in Eqs. \ref{eq6}-\ref{eq11} except that the stress variable in these equations is considered to be the effective stress.

\subsection{Dimensionless damage-elastoplasticity formulation}

Machine learning models perform best when the training data is dimensionless and has roughly a normal distribution. This is particularly important because of the way in which the network parameters are initialized and the networks are trained using a first order optimization method belonging to the gradient descent family. On the other hand, stress-strain datasets have significant differences in units and other of magnitude, with one ranging in mega or giga Pascals and the other ranging in a few percents. Therefore, it is important to work with dimensionless form of these relations that limit the range of all quantities to unity. To this end, let us introduce the scaling parameters $\sigma^*$ and $\varepsilon^*$ as some fixed values, pre-evaluated on the data. For instance, they can be the absolute maximum values of stress and strain data, respectively. From these, we can also define the elastic modulus scaling factor $E^*=\sigma^*/\varepsilon^*$. The dimensionless stress, strain, and elastic stiffness operator are then defined as 

\begin{equation}\label{eq14}
    \bar{\bs{\sigma}} = \frac{\bs{\sigma}}{\sigma^*}, \quad
    \bar{\bs{\varepsilon}} = \frac{\bs{\varepsilon}}{\varepsilon^*}, \quad
    \bar{\bs{\mc{C}}} = \frac{\bs{\mc{C}}}{E^*},
\end{equation}

\noindent
where the variables with overbars indicate their dimensionless (scaled) forms. Defining $\bar{\gamma} = \gamma / \varepsilon^*$, $\bar{\mc{F}} = \mc{F}/\sigma^*$, $\bar{\mc{G}} = \mc{G}/\sigma^*$, $\bar{\bs{q}}=\bs{q}/\sigma^*$, and $\bar{\bs{h}} = \bs{h}/E^*$, we can then express the dimensionless elastoplasticity relations as 

\begin{align}
    \dot{\bar{\bs{\sigma}}} &= \bar{\mc{C}}:(\dot{\bar{\bs{\varepsilon}}} - \dot{\bar{\bs{\varepsilon}}}^p), \label{eq15} \\
    \dot{\bar{\bs{\varepsilon}}}^p &= \dot{\bar{\gamma}} \bar{\bs{r}}(\bar{\bs{\sigma}}, \bar{\bs{q}}), \label{eq16} \\
    \bar{\bs{r}} &= \frac{\partial \bar{\mc{G}}}{\partial\bar{\bs{\sigma}}} = \frac{\partial \mc{G}}{\partial\bs{\sigma}}=\bs{r}, \label{eq17} \\
    \bar{\bs{n}} &= \frac{\partial \bar{\mc{F}}}{\partial\bar{\bs{\sigma}}} = \frac{\partial \mc{F}}{\partial\bs{\sigma}}=\bs{n}, \label{eq18} \\
    \dot{\bar{\bs{q}}} &= \dot{\bar{\gamma}} \bar{\bs{h}}(\bar{\bs{\sigma}}, \bar{\bs{q}}). \label{eq19} 
\end{align}

\noindent
Considering that the damage variable $\omega$ ranges between $(0, 1)$, there is no scaling needed for this parameter. We can therefore have the dimensionless damage-elastoplasticity relations as 

\begin{align}
    \bar{\bs{\sigma}} = (1-\omega)\bar{\bs{\mc{C}}}:(\bar{\bs{\varepsilon}} - \bar{\bs{\varepsilon}}^p). \label{eq20} 
\end{align}

\noindent
It can be observed that the dimensionless relations take almost the same form as their original version before the introduction of dimensionless parameters. 
\paragraph{Remark} For the rest of the paper, we only use these dimensionless forms and therefore drop the overbar notation for convenience. 

\subsection{Discussions}
These relations result in a set of ordinary differential equations (ODE) describing the mechanical response of solids and geomaterials. For uniaxial or multi-axial loading experiments on materials, we obtain the total nonlinear stress-stain curves at different stages of loading. The objective is then to characterize the response of the material based on a nonlinear material model. These parameters are often identified using a separate optimization loop. The challenge is that the optimization algorithm does not have access to the gradient of the loss function with respect to the parameters of the ODE system. Physics-informed neural networks, on the other hand, provide a unified approach for the solution and identification of ordinary and partial differential equations. Therefore, they can be used here effectively to characterize experiments and discover new formulations that are not easily found using classical models. We review this approach in the next section.

\section{PINN elastoplasticity solver}

As discussed, the REV elastoplasticity and damage \cref{eq14,eq15,eq16,eq17,eq18,eq19,eq20} result in a set of nonlinear ODEs, with the plastic multiplier $\gamma$ as the main unknown of the system. Therefore, given initial conditions and characteristic parameters of a material, we can solve this system of ODEs using any standard ODE solver such as the Euler's methods. Instead, we can also leverage the approximation capability of neural networks and use physics-informed neural networks to solve these system of ODEs. The main advantage is that it can be then used, with minimal changes, as an inverse solver of REV constitutive relations. In what follows, we present our proposed methodology in general, and later in the results section, we apply the proposed method for a few commonly used material laws. 

PINN ODE solvers have three main building blocks, including:
\begin{enumerate}
\item The use of neural networks to  approximate unknown variables; 
\item The use of automatic differentiation to evaluate ODE residuals; 
\item Optimization of a composite loss function, consisting of initial conditions, ODE residuals, and data (optional) \cite{Raissi2019}. 
\end{enumerate}
For the elastoplasticity problem discussed above, the variable $\gamma(t)$ is the main unknown of the problem, and other variables can be fully defined in terms of $\gamma(t)$. Therefore, instead of the  \emph{discretization} step of classical methods, $\gamma(t)$ can be approximated using a continuous, fully connected feed-forward neural network, expressed mathematically as

\begin{equation}\label{eq21}
    \gamma(t) \approx \hat{\gamma}(t) = \Sigma^L \circ \Sigma^{L-1} \circ \dots \circ \Sigma^1(t).
\end{equation}

\noindent
Here, $t$ is the input or independent variable, $\hat{\gamma}$ is the final output (approximate solution), and $\circ$ is the composition operator. The nonlinear transformation operator $\Sigma$ is expressed as

\begin{equation}\label{eq22}
\hat{\mathbf{y}}^l = \Sigma^l(\hat{\mathbf{x}}^l) :=  \sigma^l(\mathbf{W}^i\hat{\mathbf{x}}^{l-1} + \mathbf{b}^l)
\end{equation}

\noindent
where, $\hat{\mathbf{x}}^{l-1}$ is the input to, and $\hat{\mathbf{y}}^l$ is the output of any hidden layer $l$, with $\hat{\mathbf{x}}^{0}=t$ and $\hat{\mathbf{y}}^L=\hat{\gamma}$. The parameters (DOFs) of each layer are collected in weight and bias matrices $\mathbf{W}^l, \mathbf{b}^l$, and the set of all weight and bias matrices are collected conveniently in a vector $\bs{\theta}\in\mathbb{R}^D$, where $D$ is the total number of parameters of approximation \cref{eq21}. For PINNs, the nonlinear function $\sigma^l$ is commonly taken as the \emph{hyperbolic-tangent} function for layers $l=1,\dots,L-1$ and \emph{linear} for the last output layer. For convenience, we can re-write \cref{eq21} as

\begin{equation*}
    \hat{\gamma}(t) = \mc{N}(t; \bs{\theta})
\end{equation*}

\noindent
with $\mc{N}$ representing the multi-layer transformations defined in \cref{eq21,eq22}. 
Since \cref{eq21} forms a continuous function, one can use the \emph{automatic-differentiation} (AD) algorithm \cite{Gune2018}, readily available in modern deep-learning platforms such as TensorFlow, to evaluate variables such as $\dot{\gamma}$ and therefore the ODE residuals.

The last step to fully define the optimization problem is the construction of the composite (total) loss function, as 

\begin{equation}\label{eq23}
    \mc{L}(\mathbf{T}, \mathbf{L}_1, \mathbf{L_2}, \dots; \bs{\theta}) = \lambda_1 \mc{L}_1(\mathbf{T}, \mathbf{L}_1; \bs{\theta}) + \lambda_2 \mc{L}_2(\mathbf{T}, \mathbf{L}_2; \bs{\theta}) + \dots,
\end{equation}

\noindent
where $\mathbf{T}$ is the set of temporal sampling points, with $\mathbf{L}_1, \mathbf{L}_2, \dots$ as set of target (true) values for the loss terms $\mc{L}_1, \mc{L}_2, \dots$ at corresponding temporal points, with $\lambda_1, \lambda_2, \dots$ as the weight (penalty) factors for each term. Unless specified, these weights are assumed to be unity in this manuscript, i.e., $\lambda_i = 1$. Each loss term represents a constraint that should ideally be satisfied after the training. The target datasets $\mathbf{L}_i$ can be experimental data for strains or stresses or simply $\boldsymbol{0}$ for constraint residuals. The loss function of choice is the \emph{mean-squared-error} regression loss, indicated here using the norm symbol, i.e., $\left\| \circ \right\| \equiv \text{MSE}(\circ)$. 

For the constitutive theory expressed above, we deal with both equality and inequality terms. Given a function $f(t)$ and a dataset $\mathbf{T}, \mathbf{F}$, containing $N$ data-points $t_n, F_n$, representing expected values of $f$ at discrete points $t_n$, an equality loss term can be written as   

\begin{align}
\mc{L}(\mathbf{T}, \mathbf{F}; \bs{\theta}) = \left\| f(\mathbf{T}) = \mathbf{F} \right\| &= \sum_{n=1}^N \frac{1}{N}  \bigg ( f(t_n; \bs{\theta}) - F_n \bigg )^2. \label{eq24} 
\end{align}

\noindent
This loss measures the error of $f$ at any point $t_n$, and therefore, if $f$ coincides with $F_n$ at all $t_n$ points, the total loss (error) would be zero. Imposing inequality losses is more challenging. Here, we adopt the following relations

\begin{align}
\mc{L}(\mathbf{T}, \mathbf{F}; \bs{\theta}) = \left\| f(\mathbf{T}) \le \mathbf{F} \right\| &= \sum_{n=1}^N \frac{1}{N} \bigg ( S(f(t_n; \bs{\theta}) - F_n) (f(t_n; \bs{\theta}) - F_n) \bigg )^2,  \label{eq25} \\
\mc{L}(\mathbf{T}, \mathbf{F}; \bs{\theta}) = \left\| f(\mathbf{T}) \ge \mathbf{F} \right\| &= \sum_{n=1}^N \frac{1}{N} \bigg ( S(F_n-f(t_n; \bs{\theta}) ) (f(t_n; \bs{\theta}) - F_n) \bigg )^2. \label{eq26}
\end{align}

\noindent
The Heaviside step function $S$ in \cref{eq25,eq26} is defined as 
\begin{equation*}
S(f(t_n) - F_n) = 
\left\{\begin{matrix}
1 & \text{when} & f(t_n) - F_n \ge 0\\ 
0 & \text{when} & f(t_n) - F_n < 0
\end{matrix}\right.   
\end{equation*}

\begin{figure}[H]
    \centering
    \includegraphics[width=.5\textwidth]{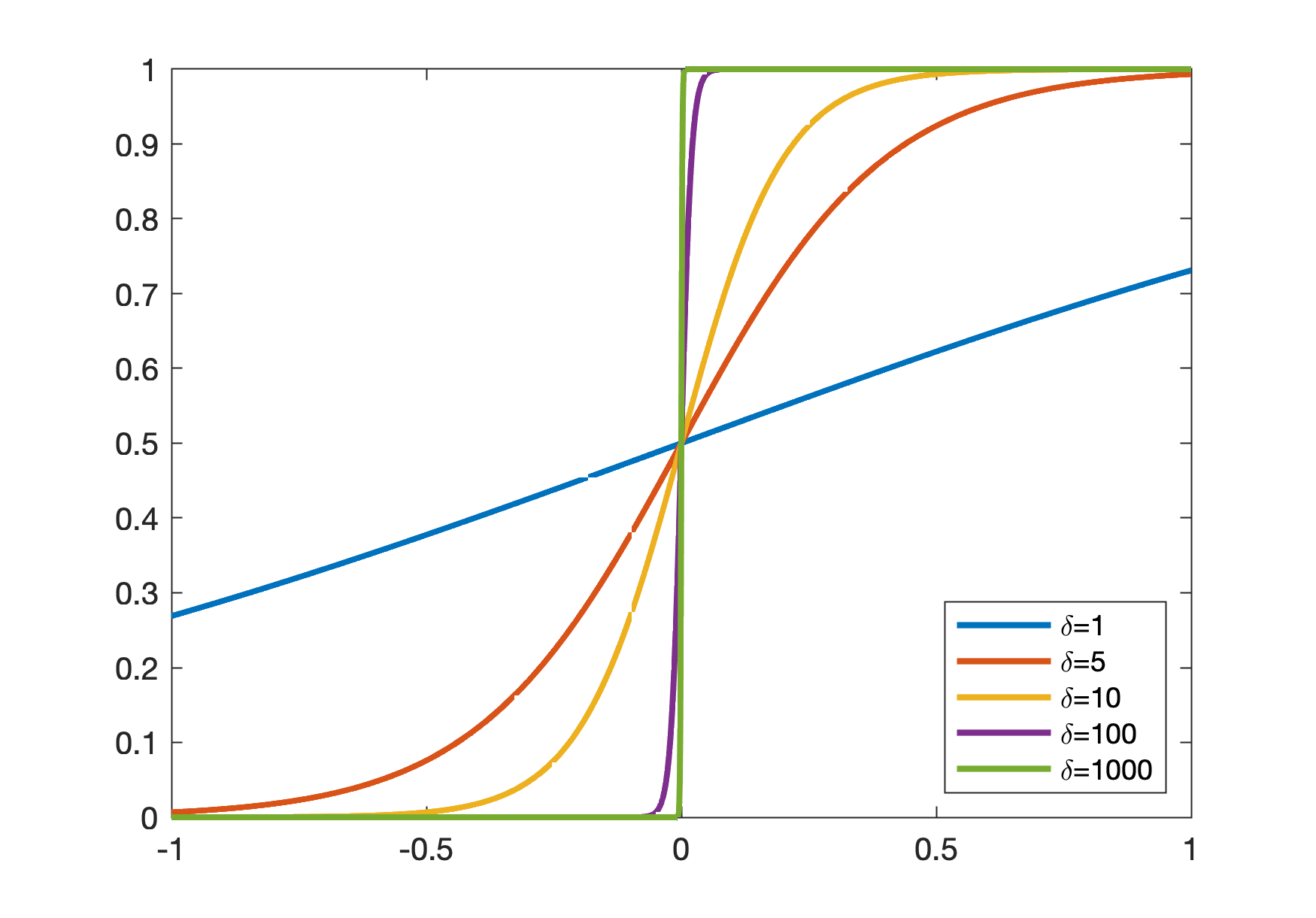}
    \caption{\emph{Sigmoid} function, evaluated on normal data, for different values of the \emph{softness} parameter $\delta$. }
    \label{fig2}
\end{figure}

\noindent
\cref{eq25} implies that at points where the inequality constraint $f(\mathbf{T}) \le \mathbf{F}$ is not met, the Heaviside function activates the loss term and imposes the lower-bound equality constraint $f(t_n) = F_n$ at these points. A similar analogy can be used to interpret \cref{eq26}. Note that for the purpose of optimization, the Heaviside step function is not a suitable choice as it does not have gradients. Therefore, for implementation purposes, we adopt the \emph{Sigmoid} function instead, which is defined as:

\begin{equation}\label{eq27}
    S(x; \delta) = \frac{1}{1 + e^{-\delta x}},  
\end{equation}

\noindent
where $\delta$ defines the \emph{softness} or \emph{hardness} of the inequality logic, as shown in  \Cref{fig2}. Given the discussion above, the constraints that should be satisfied for a PINN-elastoplasticity solver are summarized in \Cref{table1}. It should be noted that if a damage elastoplasticiy constitutive model is used, the Cauchy stress in these constraints will be replaced by effective stress and the the additional constraints associated with damage variable should be used.

\begin{sidewaystable}
\centering
\caption{Summary of constraints used to define PINN-elastoplasticity solver.}
\begin{tabular}{| m{0.05\textwidth} | m{0.13\textwidth} | m{0.10\textwidth} | m{0.3\textwidth} | m{0.30\textwidth}|}
\hline
\multicolumn{1}{|l|}{Type} & \multicolumn{1}{ c|}{Constraint} & \multicolumn{1}{ c|}{Conditions} & \multicolumn{1}{ c|}{Implementation ($\text{MSE}~\circ$)} & \multicolumn{1}{ c|}{Description} \\ \hline \hline
\multirow{2}{*}{Data} & $\sig = \mathbf{L}_{\sig}$ &   &  $\sig - \mathbf{L}_{\sig}$ & Data-driven constraints \\ \cline{2-5}
& $\eps = \mathbf{L}_{\eps}$ &   &  $\eps - \mathbf{L}_{\eps}$ & Data-driven constraints \\ \hline
\multirow{9}{*}{PINN} & $\mc{F} \le 0$ &  &  $S(\mc{F})\mc{F}$ & Non-positivity constraint of the yield function (\cref{eq9}) \\ \cline{2-5}
& $\dot{\gamma} \ge 0$ &  &  $S(-\dot{\gamma}) \dot{\gamma}$ & Non-negativity constraint of the plastic multiplier  (\cref{eq9}) \\ \cline{2-5}
& $\mc{F}\dot{\gamma} = 0$ &  & $\mc{F}\dot{\gamma}$ & KKT condition (\cref{eq9}) \\ \cline{2-5}
& $\dsig = \Ce: \deps$ & $\bs{n}:\dsig < 0$ & $S(\bs{n}:\dsig)(\dsig - \Ce:\deps)$ & Elastic Unloading \\ \cline{2-5}
& $\dsig = \Ce: \deps$ & $\bs{n}:\dsig > 0$ \newline $~\mc{F}<0$ & $S(-\bs{n}:\dsig)S(-\mc{F})(\dsig - \Ce:\deps)$ & Elastic loading \\ \cline{2-5}
& $\dsig = \Ce: (\deps - \deps^p)$ & $\bs{n}:\dsig > 0$ \newline $\mc{F}=0$ & $S(-\bs{n}:\dsig)S(\mc{F})(\dsig - \Ce:\deps - \Ce:\deps^p)$ & Elasto-plastic loading \\ \cline{2-5}
& $\dg = \frac{\bs{n}:\Ce:\deps}{\bs{n}:\Ce:\bs{r} - \partial_{\bs{q}} \mc{F}\cdot\bs{h}}$ & $\bs{n}:\dsig>0$ \newline $\mc{F} = 0$ & $S(-\bs{n}:\dsig)S(\mc{F})(\dg - \frac{\bs{n}:\Ce:\deps}{\bs{n}:\Ce:\bs{r} - \partial_{\bs{q}} \mc{F}\cdot\bs{h}})$ & Elasto-plastic loading \\ \hline
\multirow{2}{*}{PINN$^*$} & $\omega \ge 0$  &  &  $S(-\omega) \omega $ & Non-negativity constraint of the damage variable  \\ \cline{2-5}
& $\omega \le 1$ &  &  $S(\omega-1) (\omega-1) $ & Failure occurs at damage variable equal to 1  \\ \hline

\end{tabular}
\label{table1}

\raggedright
\vspace{10pt}
{~~*Applies only for coupled damage-elastoplaticity constitutive model. }

\end{sidewaystable}

Given the network parameters $\bs{\theta}\in\mathbb{R}^D$, the total loss function \cref{eq23}, and training dataset $\mathbf{T}, \mathbf{L}_1, \mathbf{L}_2, \dots$ containing $N$ training points $(t_n, (L_1)_n, (L_2)_n, \dots)$, we can express the optimization problem as

\begin{align}\label{eq28}
\bs{\theta}^* = \argmin_{\bs{\theta} \in \mathbb{R}^D} \mc{L}(\mathbf{T}, \mathbf{L}_1, \mathbf{L}_2, \dots; ~\bs{\theta}).
\end{align}

\noindent
Note that for ODE residuals or inequality constraints of elastoplasticity, the true value for that loss term is simply zero for any time step. Solving this optimization problem results in optimal values $\bs{\theta}^*$ for the problem under investigation.

\section{Results}
The von Mises elastoplasticity model and its variants such as the pressure-dependent Drucker-Prager model still remain the most commonly used constitutive model describing mechanical response of solids. Main reasons for its applicability include its thermodynamical foundation, explainable formulation and its numerical stability. Here, we apply the proposed formulation for characterization of different variations of the von Mises models with isotropic and kinematic hardening, damage, and pressure dependency. However, these applications cover the required formulation and implementation for a wider ranger of elastoplasticity models such as Cam-Clay family of models employed in soil mechanics. 

\subsection{von Mises elastoplasticity and damage models}
In its general form, the yield and potential surfaces of the isotropic von Mises model considering pressure dependency  (Drucker-Prager model) is expressed as 

\begin{align}
\mc{F} &= \mc{R} \tau - \mc{M} p - K(\alpha), \label{eq29} \\
\mc{G} &= \mc{R} \tau, \label{eq30}
\end{align}

\noindent
where $\tau$ is the equivalent shear (Mises) stress, defined as $\tau = \sqrt{3/2} \left\| \bs{\eta} \right\|$ with $\bs{\eta}$ as the deviatoric part of the kinematic stress tensor, i.e., $\bs{\eta} = \sig_d - \bs{\beta}$, and $p$ is the average confining stress, expressed as $p = -tr(\sig)/3$. 
$\mc{R}$ is the \emph{shape} parameter which controls the shape of the yield surface (see \Cref{fig1}). $\mc{M}$ is the Drucker-Prager's pressure dependency ratio. For $\mc{M}=0$ and $\mc{R}=1$, this general form yields the von Mises elastoplasticity model. 
The parameter $\alpha$ is the isotropic hardening parameter, defined as $\alpha \equiv \sqrt{2/3}~\left\|\epsd^p\right\|$. The isotropic hardening model $K(\alpha)$ is expressed as $K(\alpha) = \sigma_{Y0} + \bar{K}\alpha$, which controls the expansion of the yield surface as a function of plastic strain. This form can be general and nonlinear; for instance, we can add a quadratic term, i.e., $K(\alpha) = \sigma_{Y0} + \bar{K}\alpha +  \bar{K}_2\alpha^2$, as we later use in the case of discovery. 
The kinematic hardening tensor $\bs{\beta}$ controls the evolution of the center of the yield surface, and is defined based on the kinematic hardening rule $\dot{\bs{\beta}} = \frac{2}{3} H'(\alpha) ~\depsd^p$. Here, we assume a linear kinematic hardening model, i.e., $H'(\alpha) = \bar{H}$.  
Different aspect of the yield function is plotted in \Cref{fig1}.
Noting that 

\begin{equation*}
\bs{r} = \frac{\partial \mc{G}}{\partial \sig} = \mc{R}  \sqrt{\frac{3}{2}} \frac{\bs{\eta}}{\left\| \bs{\eta} \right\|}, \quad 
\bs{n} = \frac{\partial \mc{F}}{\partial \sig} = \bs{r} - \mc{M} \frac{1}{3}\bs{1}, \quad 
\frac{\partial \mc{F}}{\partial \bs{\beta}} = -\bs{r}, 
\end{equation*}

\noindent
then imposing the consistency condition \cref{eq10} implies that during the plastic loading ($\mc{F}=0$, $\dot{\gamma}>0$),  

\begin{equation}\label{eq31}
    \dg = \frac{\bs{n}:\Ce:\deps}{\bs{n}:\Ce:\bs{r} + (2/3~\bs{r}:\bs{r})^{1/2} K'(\alpha) + (2/3~ \bs{r}:\bs{r})H'(\alpha)}. 
\end{equation}

\begin{figure}[t]
    \centering
    \includegraphics[width=1\textwidth]{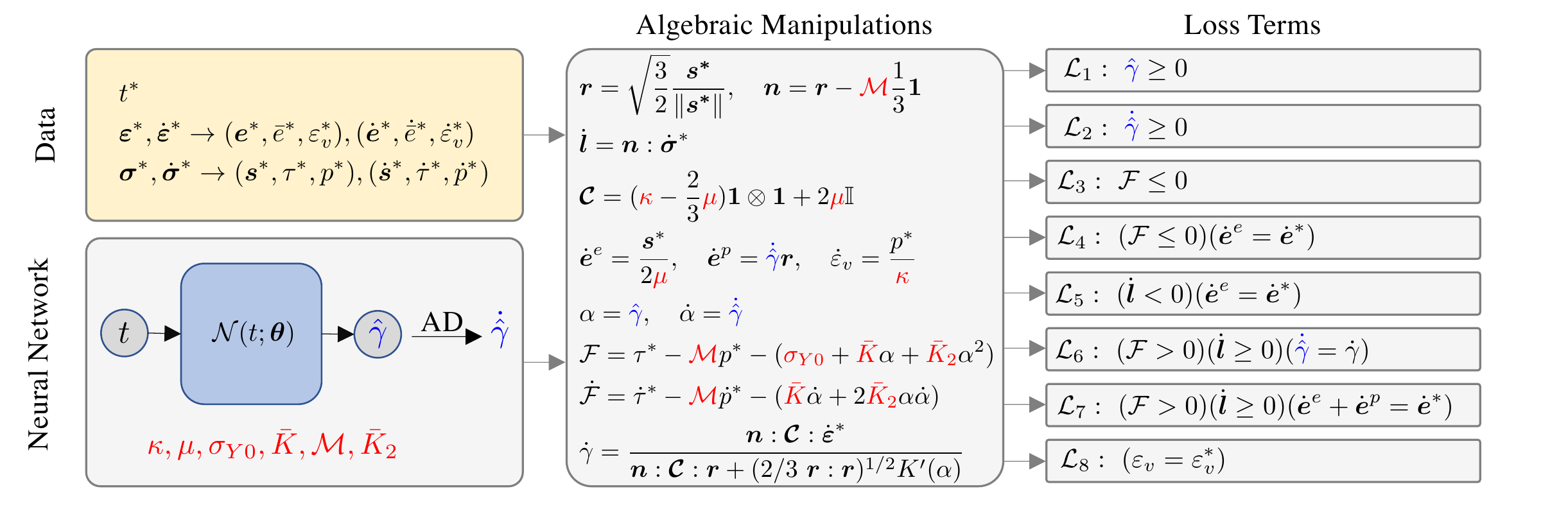}
    \caption{The proposed PINN-Plasticity solver for characterization and discovery of constitutive models. The yellow box includes the training data, i.e., strain and stress history as a function of time and their deviatoric, volumetric, and time-derivative (rate) transformations. The neural network $\mc{N}(t; \bs{\theta})$ is used to approximate the solution variable $\gamma$, i.e., the plastic multiplier (approximations are in blue). The characteristic parameters of the constitutive model are defined as trainable parameters and are presented in red color. The center box highlights the algebraic manipulations that are needed to express elastoplasticity relations. The right column highlights elastoplasticity loss terms. Note: this algorithm is valid for the isotropic hardening model. For kinematic hardening, $\bs{\beta}$ needs also to be approximated using a neural network, however, the rest of the formation remains similar. Additionally note that for the case of coupled damage-plasticity, stresses are all effective stresses.  }
    \label{fig3}
\end{figure}

\noindent
Note that for the von Mises family, we have $tr(\bs{r})=0$ and therefore $\eps^p = \eps_d^p$. For the case of isotropic damage, as described earlier, the overall form of the formulations remains intact when transformed to the effective stress space (\cref{eq12}), therefore the yield and potential functions \cref{eq29,eq30} are modified as 

\begin{align}
\mc{F} &= \frac{\mc{R} \tau - \mc{M}p}{1-\omega(\alpha)} - K(\alpha), \label{eq32} \\
\mc{G} &= \frac{\mc{R} \tau}{1-\omega(\alpha)}, \label{eq33}
\end{align}

\noindent
and the isotropic damage function $\omega(\alpha)$ is considered as $\omega = (\alpha - \alpha_i)/(\alpha_s-\alpha_i)$ for $\alpha \in [\alpha_i, \alpha_s]$, where $\alpha_i$ and $\alpha_s$ are the equivalent plastic strains at damage initiation and damage saturation, respectively. Assuming that the onset of damage and plastic deformation are the same, we have $\alpha_i=0$.
In summary, the parameters of the model include the elasticity parameters $\mu$ and $\kappa$ as shear and bulk modulus, respectively, the initial yield stress $\sigma_{Y0}$, the isotropic hardening parameter $\bar{K}$, the kinematic hardening parameter $\bar{H}$, the Drucker-Prager's pressure dependency parameter $\mc{M}$, and the damage saturation strain $\alpha_s$. The proposed PINN-elastoplasity solver is schematically shown in \Cref{fig3}.

\subsection{Validation: Uniaxial Loading}

\begin{figure}[H]
    \centering
    \includegraphics[width=0.9\textwidth]{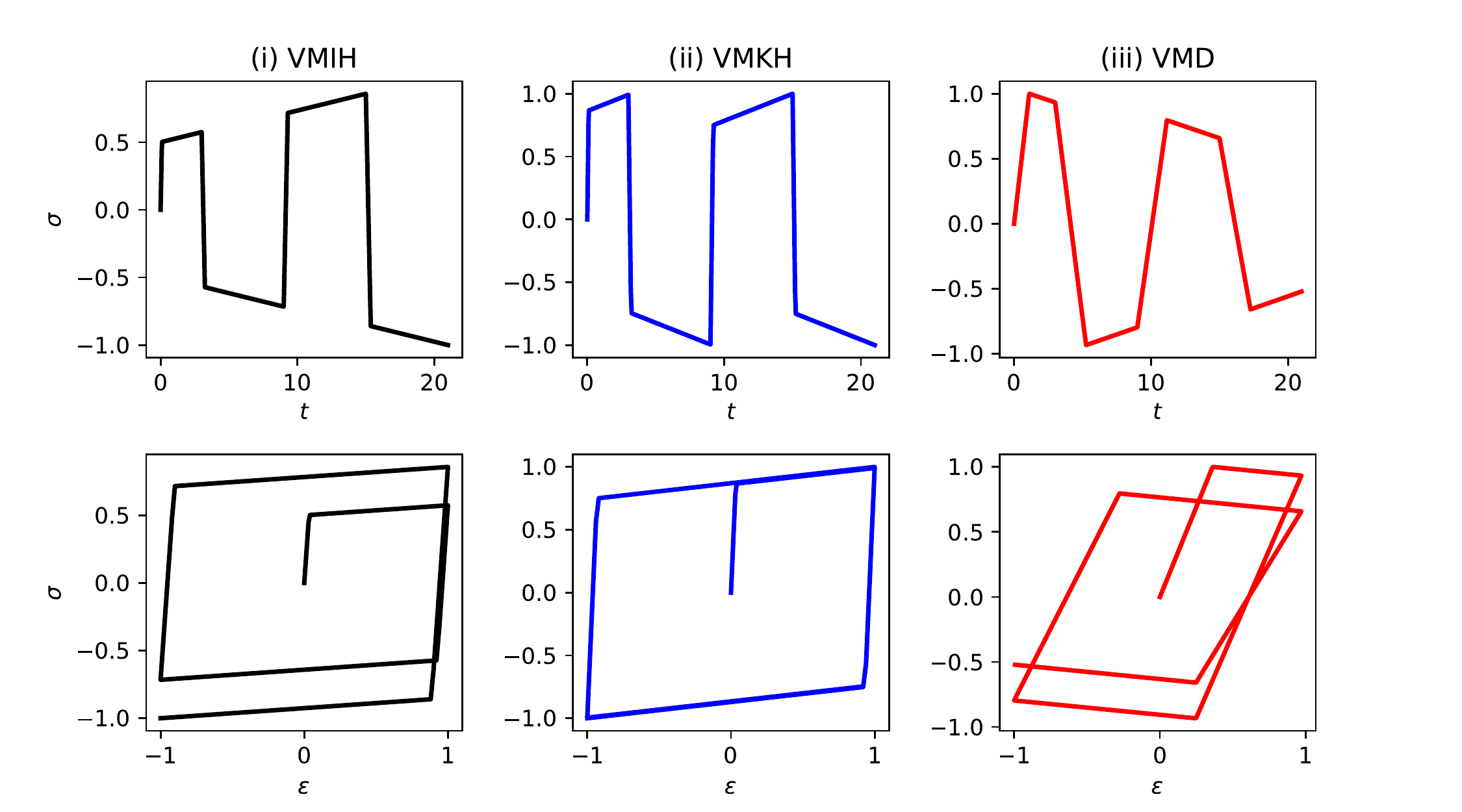}
    \caption{The dimensionless data based on the direct simulation of von Mises model with (i) isotropic hardening (VMIH), (ii) kinematic hardening (VMKH), and (iii) perfect plasticity with damage softening (VMD). The top row presents the stress history as a function of time, while the bottom row shows the stress-strain history. Isotropic hardening (i) results in expansion of the yield surface while kinematic hardening (ii) results in relocation of the yield surface. The strain softening is captured in damage-plasticity (iii).}
    \label{fig4}
\end{figure}

Let us first validate the proposed framework on three datasets generated by solving:
\begin{enumerate}[i.]
    \item The von Mises isotropic hardening (VMIH) model with parameters $\kappa=111.11 ~\text{GPa}$, $\mu=83.33~\text{GPa}$, $\sigma_{Y0}=200 ~\text{MPa}$, and $\bar{K}=10~\text{GPa}$.
    \item The von Mises kinematic hardening (VMKH) model with parameters $\kappa=111.11 ~\text{GPa}$, $\mu=83.33~\text{GPa}$, $\sigma_{Y0}=200 ~\text{MPa}$, and $\bar{H}=10~\text{GPa}$.
    \item The von Mises perfect plasticity model with damage (VMD), where parameters include $\kappa=50.2~\text{GPa}$ , $\mu=23.17~\text{GPa}$ , $\sigma_{Y0}=663~\text{MPa}$ , $\bar{K}=0~\text{GPa}$, ${\alpha}_s=0.276$. 
\end{enumerate}
The specimens are subjected to multiple strain-controlled uniaxial loading cycles, each with a constant strain-rate of $1\%$ (per unit pseudo time), shown in \Cref{fig4}. 
The datasets are generated by direct solution of elastoplasticity system of ODEs, described in previous section, using MATLAB's `ode45' solver.  
Different material responses are observed in these datasets, in which, the material's response remains linear elastic until it reaches the initial yield stress $\sigma_{Y0}$, followed by either isotropic hardening, kinematic hardening or damage softening, as depicted in \Cref{fig4}. Note that all plots are in their dimensionless (scaled) forms, with dimensionless parameters $\sigma^*, \varepsilon^*$ as the absolute maximum value of stress and strain, respectively, and $E^*$ as $E^*=\sigma^*/\varepsilon^*$. 

Given total stresses and strains applied on the specimen or obtained from the experiment, and depending on the choice of the model, the unknowns include the material parameters $(\kappa, \mu, \sigma_{Y0}, \bar{K}, \bar{H}, \alpha_s)$ and the ODE solution variable $\gamma(t)$, the kinematic hardening solution variables $\bs{\beta}(t)$, and the damage variable $\omega(t)$. The material parameters are defined as \emph{trainable} parameters for the ultimate optimization problem. The unknown solution variables $\gamma(t)$ and $\bs{\beta}(t)$ are approximated using fully-connected neural networks, i.e. $\gamma(t) = \mc{N}_\gamma(t; \bs{\theta}_{\gamma})$ and $\bs{\beta}(t) = \mc{N}_\beta(t; \bs{\theta}_{\beta})$. The damage variable $\omega(t)$ is defined as a function of the equivalent plastic strain $\alpha$ or equivalently the plastic multiplier $\gamma(t)$. 
With the aid of Automatic Differentiation, the loss terms are constructed as detailed in \Cref{table1}. The total loss function is then evaluated and optimized on the stress-strain data and constitutive relations. 

As per the network hyper parameters, we build the $\gamma(t)$ and $\bs{\beta}(t)$ neural networks with 8 hidden layers, each with 20 neurons, and with hyperbolic-tangent as the activation function of the hidden layers and the output layer with a linear activation function. 
The datasets consists of roughly 400-1000 data points, uniformly sampled in pseudo-time. Therefore, full-batch optimization is employed, with a maximum of 50,000 epochs, with exponential-decay scheduler with an initial learning rate of $10^{-3}$ decaying to $5\times 10^{-5}$ exponentially over 50,000 epochs. 
We also employ early-stopping to accelerate the training in case of no-improvements. The problem is solved using $\delta\in\left\{10, 50, 100, 200, 500\right\}$. 

\begin{figure}[H] 
    \centering
    \includegraphics[width=1\textwidth]{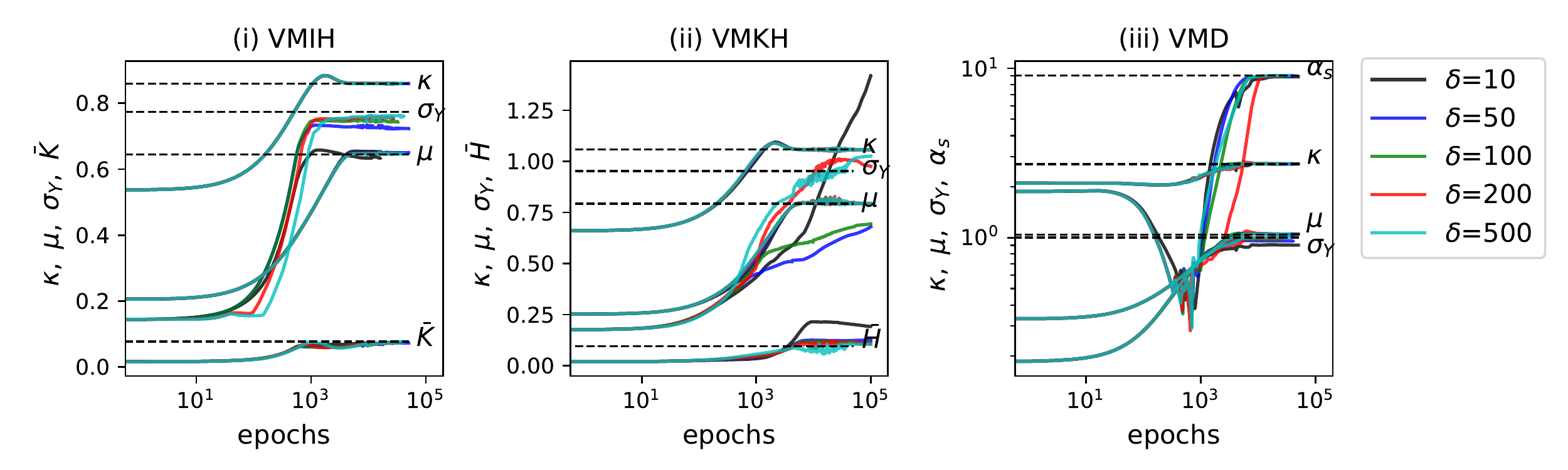}
    \caption{Constitutive characterization of the von Mises model with (i) isotropic hardening (VMIH),  (ii) kinematic Hardening (VMKH), and (iii) perfect plasticity with damage softening (VMD). The figures show the evolution of parameters as a function of epochs and for different values of $\delta$ (the softness parameter of Sigmoid function). It is found that the most accuracy is achieved for $\delta \ge 100$.}
    \label{fig5}
\end{figure}

The characterization results as a function of the training epochs and for different values of $\delta$ are plotted in Fig. \ref{fig5}.
We find that the sharper the value of $\delta$, the more accurate the results are. For $\delta \ge 100$, the framework accurately captures the parameters and the solution for the unknown $\gamma(t)$. We additionally find more sensitivity for the kinematic hardening model, and as an exception, we used more epochs (100,000) at a smaller initial learning rate ($5\times10^{-4}$) for training this problem. We associate this sensitivity with over-parameterization due to the use of separate networks for $\bs{\beta}(t)$. In general, this is unnecessary because $\bs{\beta}$ evolves as a function of $\gamma$ and normal to the yield surface, however, due to the use of AD, this can not be avoided unless a sequential training strategy is employed which we do not adopt in this study. Based on these observations, we pick $\delta=200$ for the rest of this study.

\subsection{Validation: Biaxial compression}

Since uniaxial loading (at constant strain rate) is the most commonly used experimental setup for material characterization, we primarily focus on such a loading condition to explore the proposed framework. 
However, to further demonstrate the applicability of the proposed framework to other loading conditions, we investigate the following loading scenarios that are commonly used to characterize geomaterials:
\begin{enumerate}[i.]
\item Biaxial compression (BC) at constant strain rate.
\item Undrained biaxial compression (UBC) with constant strain rate.
\item Undrained biaxial compression at variable strain rate with cyclic sinusoidal distribution (UBCS), which is commonly used for material characterization under earthquake type of loading conditions.
\end{enumerate}
The undrained loading conditions are significant in characterizing the response of geomaterials to liquefaction. Such a condition implies zero volumetric strain. As proposed in soil mechanics textbooks (see \cite{Pietruszczak2010InelasticRadius}), this loading condition is simulated by setting the lateral strain rates to one half of the axial strain rate but in the reverse direction, thereby ensuring that all components of the strain rate sum up to zero.

Most geomaterials are characterized using the Mohr-Coulomb (MC) constitutive model, which takes a hexagonal shape in $\pi$-plane that can be approximated by a smooth function as shown in \ref{fig1}. Therefore, we leverage the pressure-dependent Drucker-Prager model as a close approximation to the MC model of soil. For the biaxial setup, the sample is first loaded by a confining compressive stress $p_0$, i.e., $\sigma_1=\sigma_2=\sigma_3=p_0$, and then compressed axially while lateral loading is kept constant. For undrained loading, as discussed, a lateral strain rate is imposed to simulate the undrained condition. 

\begin{figure}[H]
    \centering
    \includegraphics[width=0.9\textwidth]{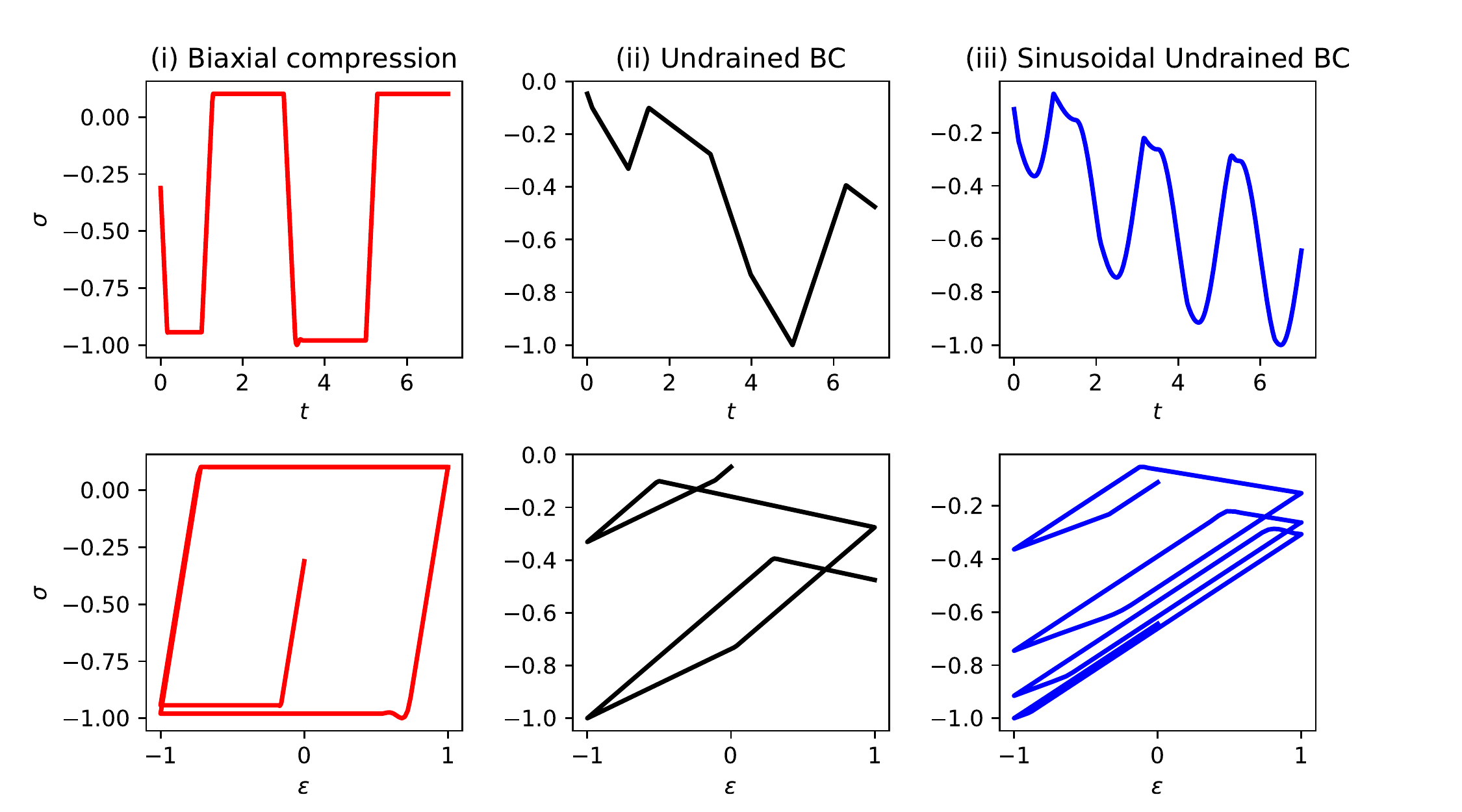}
    \caption{The dimensionless data based on the direct simulation of Drucker-Prager model, subjected to (i) cyclic biaxial compression (BC) under constant strain rate, (ii) Undraind cyclic biaxial compression (UBC) at constant strain rate, and (iii) Undraind cyclic biaxial compression (UBCS) with time-variable sinusoidal amplitude. The top row presents the stress as a function of time, while the bottom row shows the stress-strain history.}
    \label{fig6}
\end{figure}

Let us consider a representative silty soil with shear modulus, $\kappa=100~\text{MPa}$, Poisson's ratio, $\nu=0.25$, a friction angle of $25^\circ$ that is equivalent to $\mc{M}=0.466$, and a cohesive strength of $100~\text{kPa}$, i.e., $\sigma_Y=100~\text{kPa}$ and $\bar{K}=0~\text{kPa}$ for the general von-Mises model described in \cref{eq29}. The initial confining stress is set as $p_0 = -100~\text{kPa}$ (tensile positive). The amplitude of the axial strain rate for all cases is set to $0.5\%/s$, therefore the lateral strain rate for the undrained cases is set to $-0.25\%/s$.  For the sinusoidal loading, the rate of applied strain is taken as $\cos(\pi t)$. These loading conditions are shown in \Cref{fig6}. Note that due to the initial confining stress, the charts no longer start at zero. 

\begin{figure}[H] 
    \centering
    \includegraphics[width=1\textwidth]{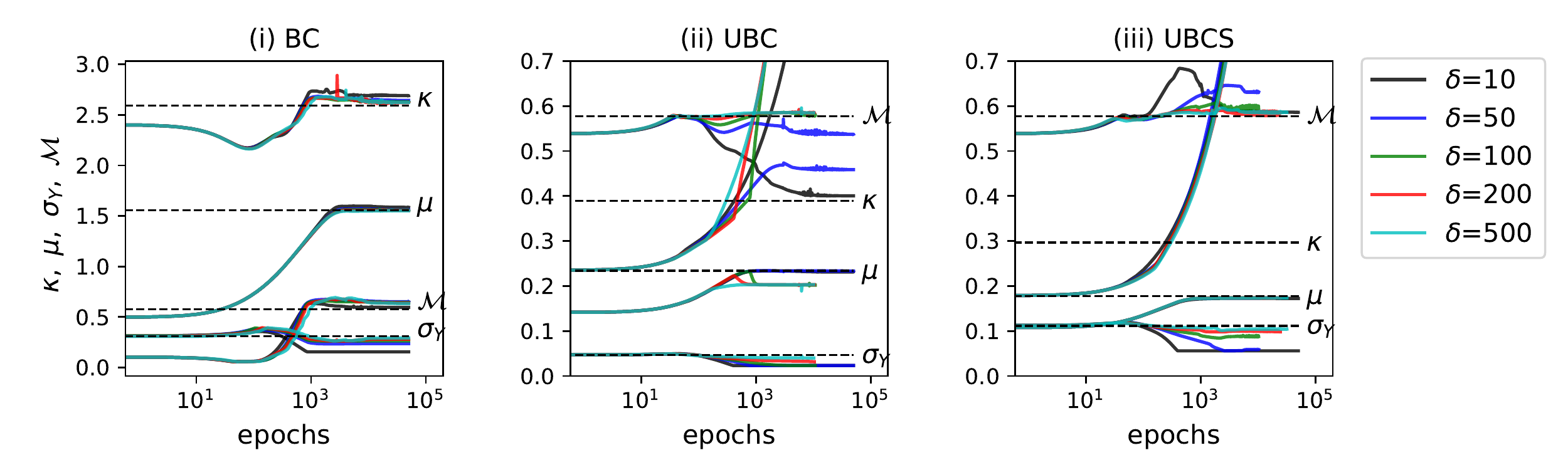}
    \caption{Constitutive characterization of the Drucker Prager model under (i) cyclic biaxial loading at constant strain rate (BC),  (ii) undrained cyclic biaxial loading at constant strain rate (UBC), and (iii) undrained cyclic biaxial loading with variable sinusoidal (UBCS) strain rate. The figures show the evolution of parameters as a function of epochs and for different values of $\delta$ (the softness parameter of Sigmoid function). Note that plots are normalized by absolute maximum stress of each data set.}
    \label{fig7}
\end{figure}

The results of parameter characterization under different loading conditions are plotted in \Cref{fig7}. For the first loading condition, i.e., biaxial compression, the framework captures all the parameters. For both undrained loading conditions, i.e., UBC and UBCS, the model is able to solve for all parameters except for the bulk modulus. This is however expected since volumetric strain, which correlates strongly with the bulk modulus, is intentionally set to zero. Therefore, bulk modulus does not contribute to the losses given the loading condition and optimizer fails to capture that. 
Lastly, we observe that for $\delta \ge 100$, the framework accurately calculates the characteristic parameters of the system. Again, all results are obtained by nondimensionalizing the data with respect to absolute maximum axial stress. 

\subsection{Exploration}

\begin{figure}[H]
    \centering
    \includegraphics[width=\textwidth]{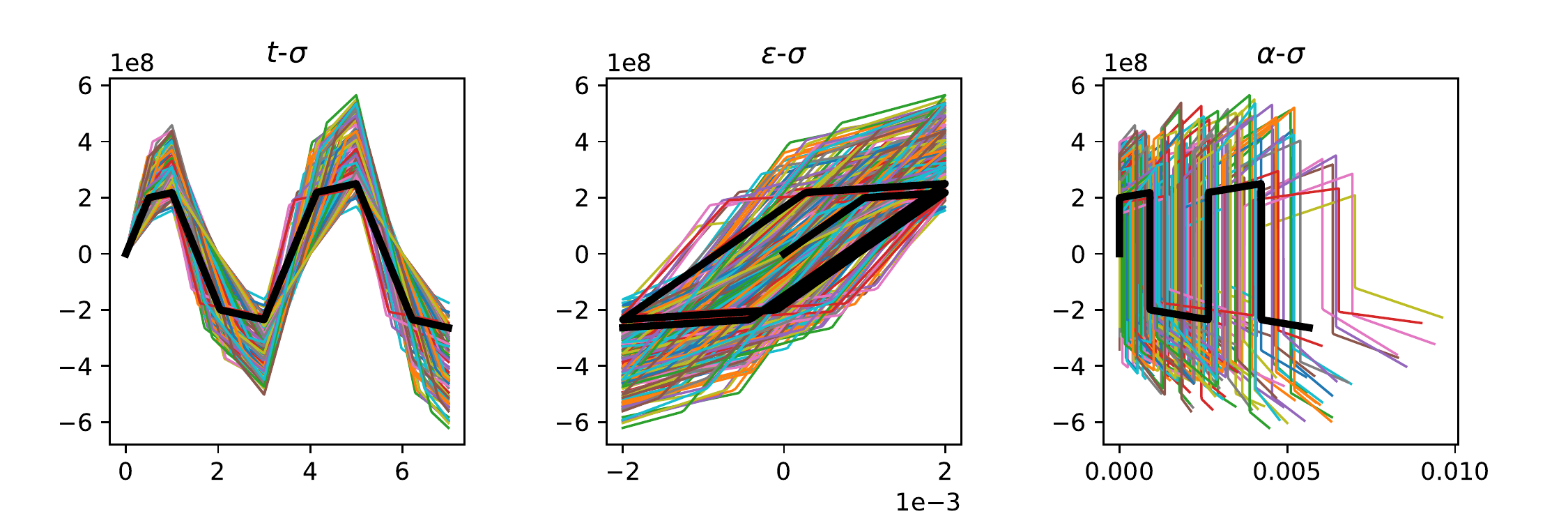}
    \caption{Training dataset with different parameters drawn independently from uniform distributions. The bold black line represent the reference training data used as the basis for transfer learning. }
    \label{fig8}
\end{figure}

Having validated the framework, we can explore its applicability and performance on a wider range of parameters and for different models. To this end, while keeping the loading unchanged, we generate 100 training datasets for VMIH, VMKH, and VMD with parameters drawn independently from uniform distributions, as summarized in \Cref{table2}, with $\mc{U}$ as the uniform distribution. These samples represent a wide range of variations in stress-strain curves and the amount of plastic deformation, as shown in \Cref{fig8} for the VMIH (case i) dataset. The solid black line in these plots is the \emph{basis-dataset} that is used to train the \emph{basis} neural network. Subsequently, the trained model is re-used (transfer learning), as the initial state, and re-trained on new samples but with only 1,000 epochs, which roughly takes 1-min using our personal laptop (with Intel i9-9980HK 8-core processor). 

\begin{table}[H]
\centering
\caption{Distribution of random parameters for building the exploration datasets.}
\begin{tabular}{ c|c c c c c c c }
\hline
    & Model & E (GPa)                & $\nu$                   & $\sigma_{Y0}$ (MPa)    & $\bar{K}$ (GPa)       & $\bar{H}$ (GPa)       & $\alpha_s$      \\ \hline
i   & VMIH  & $\mathcal{U}(100,400)$ & $\mathcal{U}(0.1, 0.4)$ & $\mathcal{U}(100,400)$ & $\mathcal{U}(1, 100)$ & -          & -                       \\ 
ii  & VMK   & $\mathcal{U}(100,400)$ & $\mathcal{U}(0.1, 0.4)$ & $\mathcal{U}(100,400)$ & -                     & $\mathcal{U}(1, 100)$          & -                       \\ 
iii & VMD   & $\mathcal{U}(40,100)$ & $\mathcal{U}(0.2, 0.4)$ & $\mathcal{U}(400,800)$ & -                     & -          & $\mathcal{U}(0.2, 0.4)$                      \\ \hline
\end{tabular}
\label{table2}
\end{table}

\begin{figure}[H]
    \centering
    \includegraphics[width=1\textwidth]{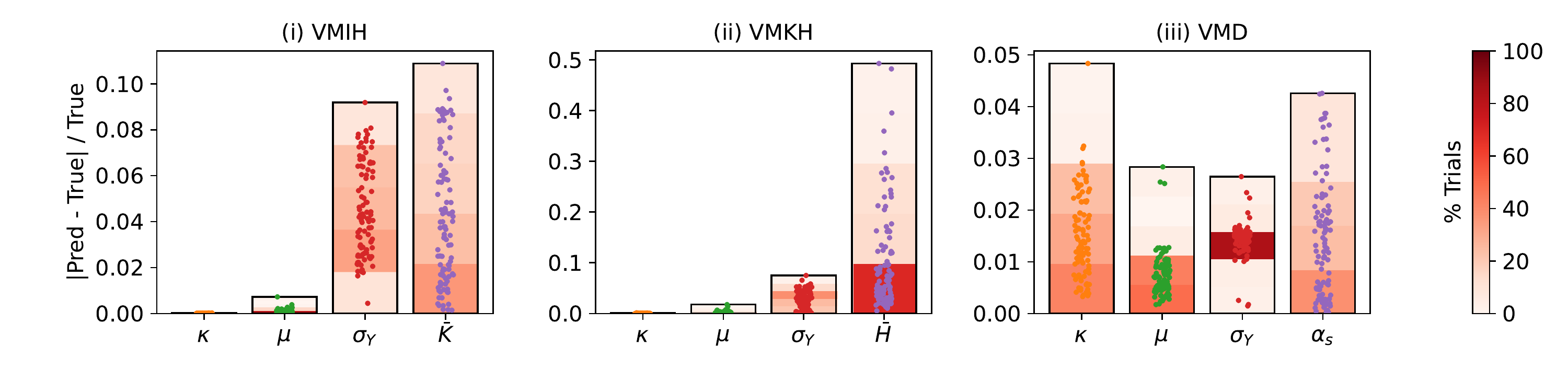}
    \caption{The relative error for each parameter identified after retraining the reference network on 100 randomly generated datasets. The re-training epochs is limited to 1,000 and takes roughly 1-min to complete. The relative error for all parameters remain very low. }
    \label{fig9}
\end{figure}

The error plots are shown in \Cref{fig9}, with cases i-iii being plotted from left to right, respectively. We find that the training works well on all parameters for majority of samples. Note that the total re-training epochs is capped at 1,000 for demonstration purposes, however, allowing more training would improve the accuracy of the identified parameters. Additionally, the maximum error occurs for the sample with the largest deviation from the basis dataset, which is somewhat expected. 

\subsection{Training epochs}
\begin{figure}[H]
    \centering
    \includegraphics[width=1\textwidth]{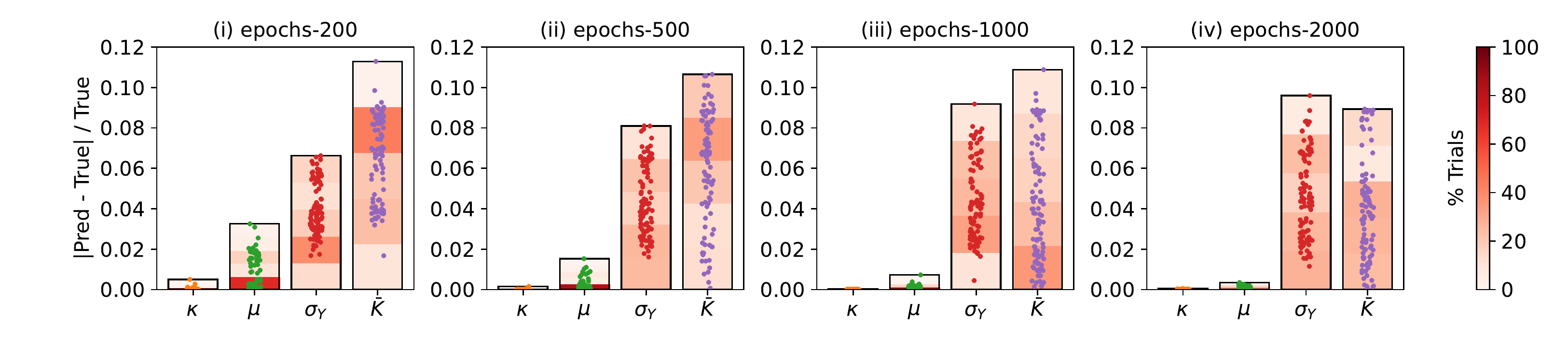}
    \caption{The relative error for VMIH trained with different number of epochs: (i) 200, (ii) 500, (iii) 1000, and (iv) 2000. Whilst increased training improves the accuracy, 1000 epochs seems to be sufficient. }
    \label{fig10}
\end{figure}

To ensure that our results are convergent and to show that 1,000 epochs provide an adequate number of training epochs for re-calibration of data, here, we present results obtained using 200, 500, 1000, and 2000 training epochs for the VMIH model. The results are shown in \cref{fig10}. It can be seen that as we increase the training time, the accuracy of the parameters improves and the distribution of error shrinks. Additionally, we observe that the results of 1000 and 2000 training iterations are very similar, therefore,  the 1000 re-training epochs seems sufficient. 

\subsection{Sample size}

\begin{figure}[H]
    \centering
    \includegraphics[width=1\textwidth]{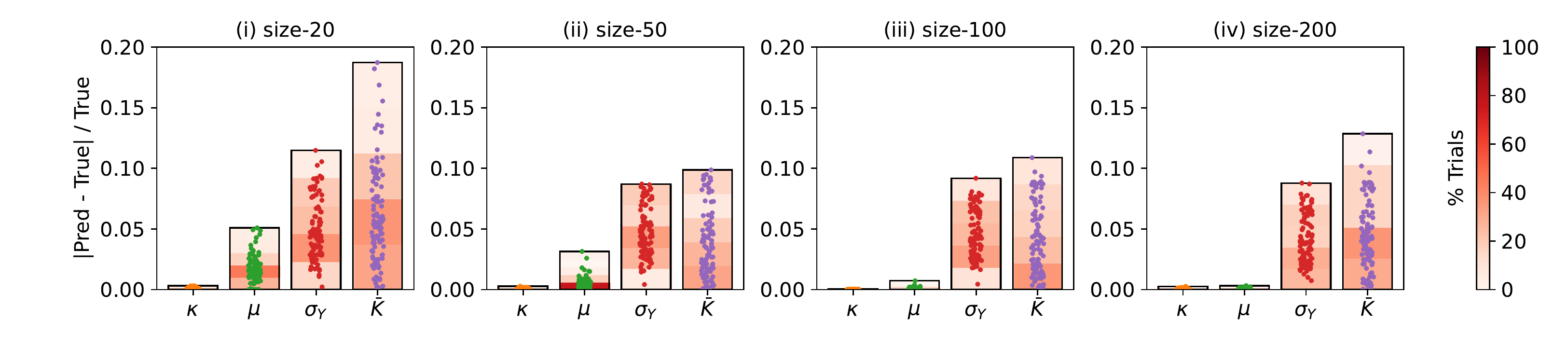}
    \caption{The relative error for VMIH with different training sample sizes, at (i) 20, (ii) 50, (iii) 100, and (iv) 200 sampling points per loading cycle. With larger sample sizes, the error distribution shrinks and the results become more accurate. }
    \label{fig11}
\end{figure}

To explore the role of sample size on our results, here we consider three additional cases to the VMIH model, with sample sizes of 20, 50, and 200 per cycle. Note that all previous cases are with sample size of 100 points per cycle. The results are plotted in \Cref{fig11}. Sample size 20 corresponds to the highest amount of error. Sample sizes 50 and 100 show almost the same distributions of error and sample size 200 seems to have slightly higher accuracy in elastic properties. Note that these are temporal sample sizes, which are resolutions that are very possible to achieve. It is often much harder to achieve spatial sample sizes.

\subsection{Sensitivity to noise}

\begin{figure}[H]
    \centering
    \includegraphics[width=1\textwidth]{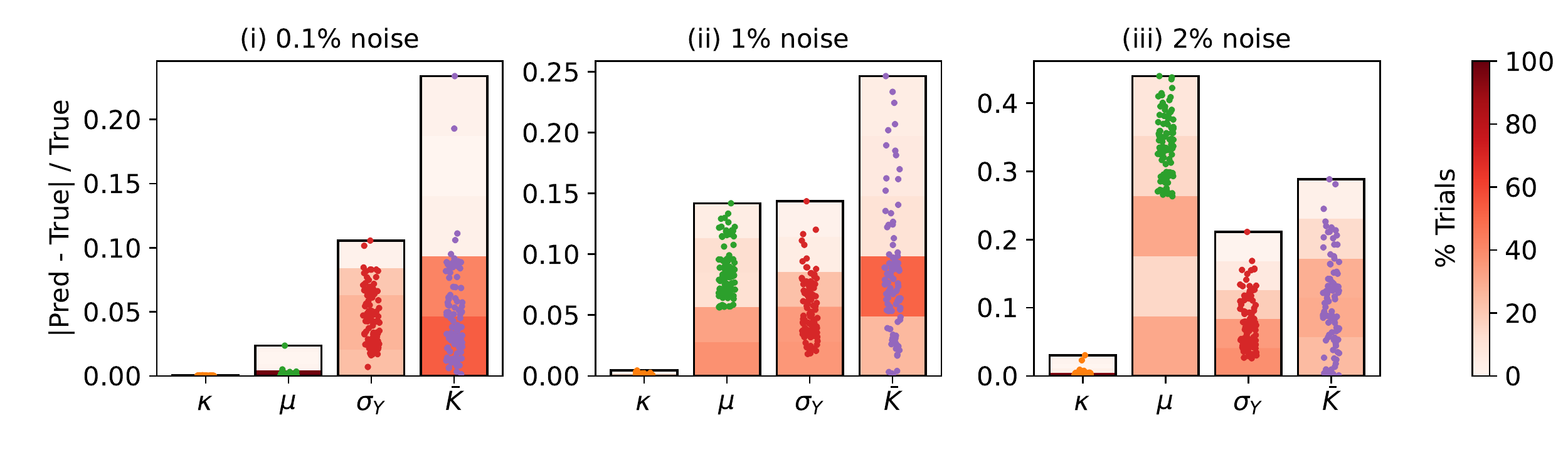}
    \caption{The relative error for VMIH with different levels of added noise, at (i) 0.1\%, (ii) 1\%, and (iii) 2\%. With added noise, the distribution of error expands and the accuracy of the parameters obtained from inverse analysis reduces. }
    \label{fig12}
\end{figure}

Since we primarily use simulated data from direct solution of elastoplasticity models, in contrast to physical (experimental) data, ours is noise-free. To simulate experimental conditions, here we re-analyze the VMIH cases with added synthetic noise. To this end, we consider adding 0.1\%, 1\%, and 2\% random noise to stress and strain data and re-perform the previous studies. Due to the added noise, we now perform each calibration with 2,000 epochs. The results are shown in \Cref{fig12}. As expected, the added noise reduces the accuracy of the parameters obtained from inverse analysis. However, the results are mostly reasonable within a small error. Shear modulus seems to have the highest sensitivity to the noise. 

\subsection{Discovery}
In the first cases, we performed full inverse analysis, i.e., without any prior assumption on parameters, under uniaxial and biaxial loading conditions. 
We then explored model calibration, meaning that we calibrated pre-trained models on new data to identify their parameters. 
In this section, we explore discovery, meaning that we add additional parameters that are irrelevant to the data, and quantify if the solver can automatically nullify those. This is often an ill-posed problem since we may arrive at cases that multiple choices can represent the same dataset. 

\begin{figure}[H]
    \centering
    \includegraphics[width=0.9\textwidth]{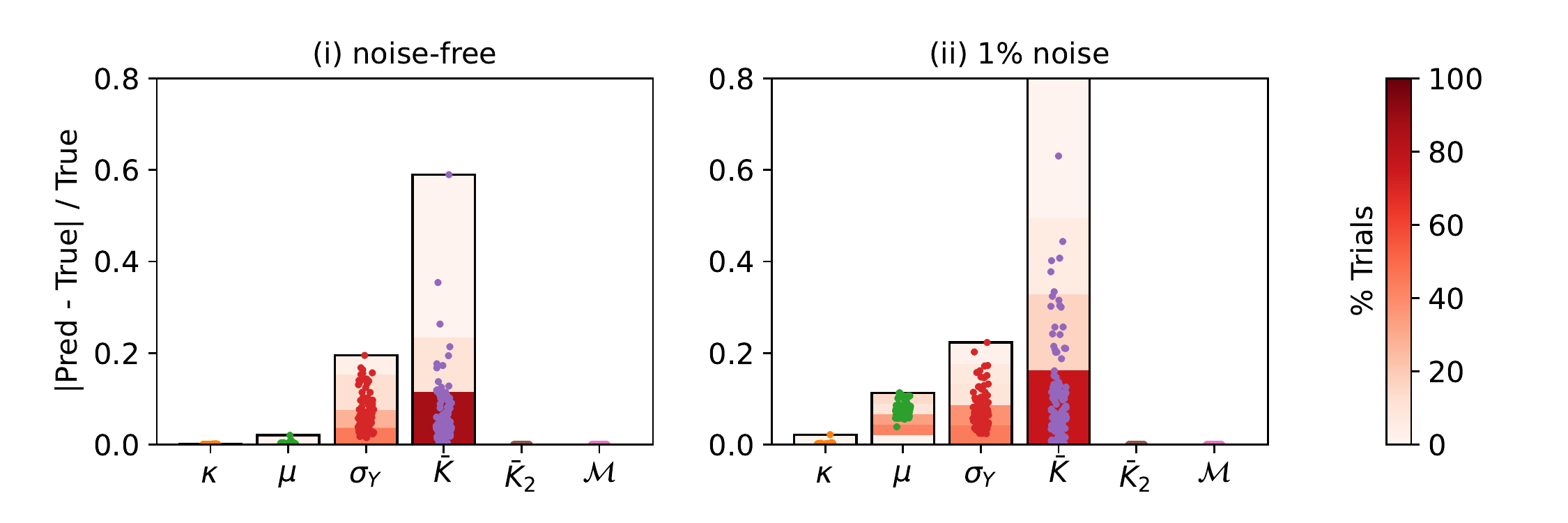}
    \caption{Performing model discovery, i.e., model that has a more general form, on the random samples generated from von Mises with isotropic hardening (VMIH) model, for (i) noise-free data and (ii) data with added 1\% Gaussian noise. The bars show the relative error for each parameter, which are captured accurately. In the case of discovery, the re-training epochs is limited to 5,000.}
    \label{fig13}
\end{figure}

In the last case, we use the dataset that was generated using the isotropic-hardening  (VMIH) model and re-train it with a more general form of the yield surface that accounts also for Drucker-Prager's pressure dependency and quadratic isotropic hardening as described in \cref{eq29}, with the shape factor $\mc{R}=1$. 
Again, we use transfer learning from the reference trained model, but this time with 5,000 re-calibration epochs. 
Additionally, we consider two cases, one with noise-free data and the other with 1\% added noise.
The results are shown in \Cref{fig13}. Once again, the model identifies the correct parameters very accurately for most of the training samples. This time, however, on very few occasions, the model cannot converge within 5,000 epochs and results show relatively large errors for parameters. We verify that using more calibration epochs and smaller learning rate, the model can learn also on those samples accurately but it is computationally less efficient. 

\section{Conclusions}
In this study, we presented a novel framework for constitutive model characterization and discovery based on the physics-informed neural networks (PINN). The framework was validated on synthetic data generated by direct solution of the plasticity ODEs and constraints. We validated the framework on uniaxial loading and biaxial loading under drained and undrained conditions, however, the formulation and implementation remains general and can be used with any loading type. We tested the framework on a wide range of material parameters and stress-strain curves and showed that the framework can efficiently and accurately recover the underlying constitutive models. Finally, we performed a general model discovery, where we added additional terms to the yield surface including pressure dependency and high-order hardening, yet the framework accurately recovered the underlying constitutive model and correctly set the additional parameters to zero. Without any special optimization on the algorithm, our model can perform model characterization in a few minutes with relatively high-precision, with most cases having errors less that 10\%.

While a recent trend in constitutive modeling suggests a general model-free formulation using complex neural network architectures, we question their applicability for large scale engineering problems due to the computational cost associated with inference from such complex networks. While such models remain extremely helpful to replace large data obtained from data-intensive experiments or from meso- or micro-mechanical or even molecular dynamic simulations, we believe that they need to be transformed to elastoplasticity-based formulations to be computationally efficient for complex, large-scale engineering analysis using the finite element method, for example. Therefore, we recognize the main advantage of our proposed framework as an explainable machine learning system for constitutive modeling and discovery of such datasets. As a next step, the authors are working on validating the framework on experimental data obtained from fracture tests on notched composite materials, which will be presented in a follow-up study. 

There are numerous paths that can be explored from this study. Here, we proposed a single architecture to capture all parameters at the same time. However, we suspect that as the constitutive model becomes more complex and with more correlated parameters, like advanced models of soil and rock mechanics, a pipe-line training strategy would work better. In other words, we can divide the training task to first identify the elastic parameters from early stages of loading or from unloading path, and then proceed to determining parameters that control early plasticity stages, and so on. Combined with sequential iterations, we believe that the proposed framework can be used to discover truly complex models. Selecting a functional form for the shape of the yield surface is also an interesting area of exploration. Hyper-elastic and hyper-plastic models are often more desirable, however, they pose challenges for their characterization as they require an energy function whose derivatives return stresses. Such a framework can be used to identify new energy-based models of materials. 

\section*{Acknowledgements}
The authors acknowledge the financial support provided by Mitacs and Natural Sciences and Engineering Research Council of Canada (NSERC).

\section*{Contributions}
E.H. designed the research, formulated and implemented the PINN-plasticity framework, and wrote the manuscript. S.A. and R.V. formulated and implemented the PINN-damage-plasticity framework, and helped with writing and revising the manuscript.

\section*{Data availability statement}
All data, models, or code generated or used during the study are available in a repository online (\href{https://github.com/sciann/sciann-applications/tree/master/SciANN-ConstitutiveModeling}{https://github.com/sciann/sciann-applications/tree/master/SciANN-ConstitutiveModeling}).

\bibliographystyle{plainnat}

\bibliography{references.bib}

\end{document}